%% file: main.tex
\documentclass[acmtog]{acmart}
\acmSubmissionID{334}

\usepackage{booktabs} 
\usepackage{multirow}
\usepackage{array}
\newcolumntype{L}[1]{>{\raggedright\let\newline\\\arraybackslash\hspace{0pt}}m{#1}}
\newcolumntype{C}[1]{>{\centering\let\newline\\\arraybackslash\hspace{0pt}}m{#1}}
\newcolumntype{R}[1]{>{\raggedleft\let\newline\\\arraybackslash\hspace{0pt}}m{#1}}

\usepackage[ruled]{algorithm2e} 

\SetAlFnt{\small}
\SetAlCapFnt{\small}
\SetAlCapNameFnt{\small}
\SetAlCapHSkip{0pt}

\setcopyright{rightsretained}
\acmJournal{TOG}
\acmYear{2021}
\acmVolume{40}
\acmNumber{6}
\acmArticle{1}
\acmMonth{12}
\acmDOI{10.1145/3478513.3480545}

\begin{document}
\title[Modeling Clothing as a Separate Layer for an Animatable Human Avatar]{Modeling Clothing as a Separate Layer for an Animatable \\ Human Avatar}

\author{Donglai Xiang}
\affiliation{%
 \institution{Carnegie Mellon University}
 \city{Pittsburgh}
 \country{USA}}
\affiliation{%
 \institution{Facebook Reality Labs Research}
 \city{Pittsburgh}
 \country{USA}}
\email{donglaix@cs.cmu.edu}

\author{Fabian Prada}
\affiliation{%
 \institution{Facebook Reality Labs Research}
 \city{Pittsburgh}
 \country{USA}}
\email{fabianprada@fb.com}

\author{Timur Bagautdinov}
\affiliation{%
 \institution{Facebook Reality Labs Research}
 \city{Pittsburgh}
 \country{USA}}
\email{timurb@fb.com}

\author{Weipeng Xu}
\affiliation{%
 \institution{Facebook Reality Labs Research}
 \city{Pittsburgh}
 \country{USA}}
\email{xuweipeng@fb.com}

\author{Yuan Dong}
\affiliation{%
 \institution{Facebook Reality Labs Research}
 \city{Pittsburgh}
 \country{USA}}
\email{ydong142857@fb.com}

\author{He Wen}
\affiliation{%
 \institution{Facebook Reality Labs Research}
 \city{Pittsburgh}
 \country{USA}}
\email{hewen@fb.com}

\author{Jessica Hodgins}
\affiliation{%
 \institution{Carnegie Mellon University}
 \city{Pittsburgh}
 \country{USA}}
\affiliation{%
 \institution{Facebook AI Research}
 \city{Pittsburgh}
 \country{USA}}
\email{jkh@cs.cmu.edu}

\author{Chenglei Wu}
\affiliation{%
 \institution{Facebook Reality Labs Research}
 \city{Pittsburgh}
 \country{USA}}
\email{chenglei@fb.com}

\renewcommand\shortauthors{Xiang, D. et al}

\begin{abstract}
We have recently seen great progress in building photorealistic animatable full-body codec avatars, but generating high-fidelity animation of clothing is still difficult. To address these difficulties, we propose a method to build an animatable clothed body avatar with an explicit representation of the clothing on the upper body from multi-view captured videos. We use a two-layer mesh representation to register each 3D scan separately with the body and clothing templates. In order to improve the photometric correspondence across different frames, texture alignment is then performed through inverse rendering of the clothing geometry and texture predicted by a variational autoencoder. We then train a new two-layer codec avatar with separate modeling of the upper clothing and the inner body layer. To learn the interaction between the body dynamics and clothing states, we use a temporal convolution network to predict the clothing latent code based on a sequence of input skeletal poses. We show photorealistic animation output for three different actors, and demonstrate the advantage of our clothed-body avatars over the single-layer avatars used in previous work. We also show the benefit of an explicit clothing model that allows the clothing texture to be edited in the animation output.
\end{abstract}



\begin{teaserfigure}
\centering
\includegraphics[width=7.05in]{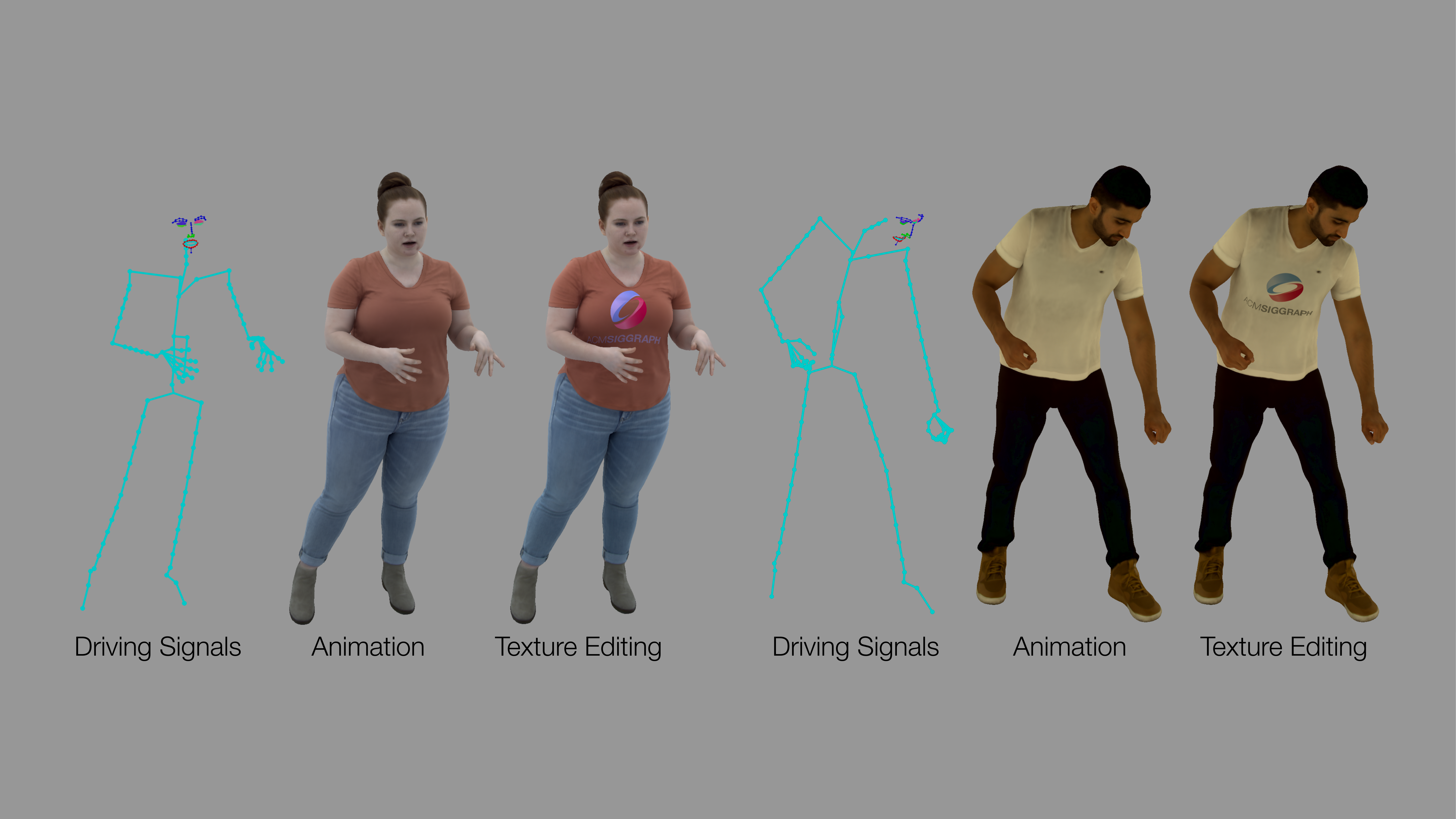}
\caption{Given a novel sequence of skeletal poses and facial keypoints as input, our proposed two-layer codec avatars produce photorealistic animation output, where the clothing texture can be consistently edited. From left to right, we show driving signals, animation output and editing results for two subjects.}
\end{teaserfigure}

\maketitle

\input{introduction}
\input{related}
\input{method}
\input{results}
\input{conclusion}

\bibliographystyle{ACM-Reference-Format}
\bibliography{bibliography}

\end{document}

%% file: introduction.tex
\section{Introduction}

Animatable photorealistic digital humans are a key capability for enabling social telepresence, and have the potential to open up a new way for people to remain connected without geographic constraints.
Early work on human body modeling built low-dimensional geometric representations of the body surface with minimal clothing~\cite{loper2015smpl,MANO2017,STAR:2020}. As a separate field of work, cloth simulation has been studied and used to create clothing deformation that does not conform tightly to the human body~\cite{baraff1998large,Kavan2011physics,narain2012adaptive,buffet2019implicit}. However, both these lines of work focus on modeling only the geometry, and cannot directly produce photorealistic rendering output. Even with the recent data-driven methods using neural networks (for example \cite{lahner2018deepwrinkles}), animating a photorealistic clothed human is still far from a solved problem.


In this work, we seek to build photorealistic full-body clothed avatars that can be animated with driving signals that can be easily accessed, for example, 3D body pose and facial keypoints. Simultaneously modeling both geometry and texture with a deep generative model, like Variational Autoencoders (VAE), has been demonstrated to be an effective way to create photorealistic face avatars~\cite{lombardi2018deep}. Recently, Bagautdinov and colleagues~\cite{bagautdinov2021driving} extend this approach to model full-body avatars with VAE, conditioned on body pose and facial keypoints. Because these conditional signals cannot uniquely describe the states for the clothing, hair and gaze, the VAE latent code is used to distinguish between these different states. In addition, it is essential to disentangle the effects of driving signals and the latent code, in order to reduce the spurious correlations between them. 

Despite the progress in previous work~\cite{bagautdinov2021driving}, challenges still remain in building high-fidelity animatable full-body avatars, and we identify the modeling of clothing as one major difficulty. Artifacts include the imperfect correlation between body pose and clothing state, ghosting effects along the boundary between clothing and skin, as well as loss of wrinkle details and dynamics in the clothing. These artifacts become more noticeable when the captured clothing is loose and the performer moves more dynamically. On the one hand, due to registration error, the network may underfit the data, making it unable to reproduce high-frequency clothing detail; on the other hand, in spite of the disentanglement, the network may still overfit, capturing unwanted chance correlation between the driving signal and the clothing state.


In this work, we explicitly represent the body and clothing as separate layers of meshes in a codec avatar. The separation leads to several benefits. First, it allows us to accurately register both body and clothing, especially with our newly developed photometric tracking approach that uses inverse rendering to align clothing texture to a reference. Second, modeling the body and clothing in separate layers alleviates the aforementioned problem of chance correlation for a single-layer avatar, as the separate layers are naturally disentangled from each other. With our two-layer VAE, a single frame of joint angles can well describe the body state, while the clothing dynamics can be inferred from the sequences of poses with a Temporal Convolutional Network (TCN), which evolves the clothing state in a way that is consistent with the body motion. Third, thanks to the explicit modeling of clothing, the animation output can be further edited by changing the clothing texture. 

To summarize, our contributions are as follows:
\begin{itemize}
    \item We present an animatable two-layer codec avatar model for photorealistic full-body telepresence; our proposed avatar can produce more temporally coherent animation with sharper boundaries and fewer ghosting artifacts compared to a single-layer avatar;
    \item Inverse rendering with our proposed two-layer codec avatar allows a photometric tracking algorithm that aligns the salient clothing texture, significantly improving correspondence in the registered clothing meshes;
    \item We demonstrate an application of our two-layer codec avatar for editing of the clothing texture that is hard to achieve with the single-layer model used in previous work.
\end{itemize}

We evaluate the proposed pipeline on the captured sequences of three different actors. We demonstrate the effectiveness of our proposed method against alternative approaches. We show that our model, with only a sequence of poses and facial keypoints as input, achieves high-quality body animation and rendering with photorealistic clothing that can be viewed from arbitrary viewpoints. 

%% file: related.tex
\section{Related Work}

Our goal in this paper is to build a realistic virtual avatar of a human that can be animated by driving signals of skeletal poses and facial keypoints to create a telepresence experience.
The \textbf{classical pipeline} for modeling such an animatable avatar typically relies on building a textured template mesh from a 3D scan and rigging the template mesh to a parameterized skeleton model such that the deformation of the template mesh is associated with the skeletal pose according to the skinning weights.
The most commonly used skinning method is the Linear Blend Skinning (LBS), which we also use to model the skeletal motion.
In the literature, many methods have been developed in order to reduce the unnatural skinning artifacts that occur with LBS, e.g.,~\cite{Kavan-08-SDQ,Kavan-05-SBS}.
However, a fundamental disadvantage of these approaches is that high-frequency deformations of skin and clothing, such as muscle bulging, folds, and wrinkles, cannot be precisely modeled.
In order to solve this problem, pose dependent blend shapes~\cite{lewis2000pose} have been proposed to reduce skinning artifacts. These blend shapes are corrective shapes that can be interpolated with respect to the pose and added to the skinned mesh.
Although blend shapes work well for skin and tight clothing, the non-rigid deformation of soft tissue and loose clothing is not modeled well by this approach.

\textbf{Physical simulation} provides an automatic way to create secondary motion of virtual characters, such as muscle bulging and cloth deformation.
Cloth simulation is typically not real-time due to the computational complexity and therefore many of the earlier methods focus on efficiency~\cite{Goldenthal2007effcient,Gillette2015realtime,Kavan2011physics,kim2013near,wang2010example}.
More recent research tackles efficiency by learning the mapping from body pose and shape to the clothing deformation produced by physical simulations~\cite{Gundogdu_2019_ICCV,Santesteban2019learning,vidaurre2020virtualtryon,Chentanez2020cloth,patel20tailornet,bertiche2020cloth3d,Jin2020pixel,bertiche2020pbns,zhang2021deep,wang2019learning}.
Among those methods, one notable concurrent work \cite{santesteban2021self} adopts a similar strategy to model clothing with a VAE and animates clothing with a temporal model. Compared with our work, this approach focuses on avoiding collision in the clothing output, but does not model clothing from real-world captured data, or produce a photo-realistic rendering of the clothing.
Cloth simulation has been leveraged in human performance capture to produce more realistic dynamic deformation of the clothing.
Stoll and colleagues reconstruct a time-varying surface geometry of the clothing from multiview video recordings and then estimate the parameters of a physical simulation model of the clothing~\cite{Stoll2010video}.
SimulCap contributes a monocular human performance capture system that not only captures the skeleton motion but also simulates cloth dynamics and cloth-body interactions~\cite{Yu2019simulcap}.

\textbf{Data-driven human modeling} has been leveraged very effectively in recent years.
The seminal work, SCAPE~\cite{SCAPE05}, learns a parametrized human body shape model from a large-scale dataset of 3D scans.
A variation of SCAPE that integrates the learned pose dependent blend shapes, SMPL~\cite{loper2015smpl}, has been widely used for human modeling and pose estimation.
However, these models can only model a human body dressed in skin tight clothing.
In order to synthesize the deformation of clothing, apart from the aforementioned simulation-based learning approaches, many methods resort to learning the deformation from real 4D capture data.
DeepWrinkle~\cite{lahner2018deepwrinkles} consists of two modules that learn the global cloth deformation in a PCA subspace as well as high frequency details, such as finer wrinkles, on a normal texture.
Similarly, Ma and colleagues learn a pose-dependent clothing shape from 4D scans with different geometric representation, including mesh-based graph convolution \cite{CAPE:CVPR:20}, surface
elements \cite{ma2021scale} and implicit functions \cite{saito2021scanimate}. Compared with our work, these methods mostly focus on modeling the clothing geometry, with less effort on creating photo-realistic rendering of clothing appearance.

Another family of generative human modeling methods does not focus on the 3D geometry, but aims to synthesize photo-realistic human images.
These neural rendering approaches typically formulate the task as an image translation problem, and learn the mapping from joint heatmaps~\cite{aberman2019deep}, rendered skeleton~\cite{Chan_2019_ICCV,si2018multistage,pumarola2018unsupervised,esser2018towards}, or rendered meshes~\cite{wang2018video,liu2019neural,liu2019liquid,Sarkar2020,prokudin2021smplpix,raj2021anr}, to real images. 
In contrast to these approaches, Deep Appearance Models~\cite{lombardi2018deep} explicitly handle \textit{both} facial geometry and appearance in the form of view-dependent texture, and is capable of producing view-dependent effects and correcting geometric artifacts. In recent work, Bagautdinov and colleagues extend deep appearance models to full bodies~\cite{bagautdinov2021driving}. However, as this method does not explicitly model clothing, it may struggle in settings where clothing is loose or exhibits significant dynamics.
Most related to our paper is the concurrent work of Habermann and colleagues~\cite{habermann2021}. This work addresses a similar problem of creating a dynamic free-view point rendering of a specific subject given skeleton motion as input. It uses a neural network to regress the clothed body shape represented by an embedded graph plus additional deformation and a dynamic texture. Compared with this work, our method uses a two-layer formulation for both registration and modeling that enables high-quality animation output.

\textbf{Dynamic scene capture} is an alternative yet less practical approach for telepresence, because it does not compress the dynamic information of the scene as a latent code like our approach, and therefore requires a much higher communication bandwidth.
That said, our method is still highly related to these methods, as we rely on dynamic scene capture to obtain training data.
Most of the existing approaches rely on multi-camera systems to recover detailed geometry using silhouettes or photometric stereo.
They reconstruct either the shapes of each individual time step~\cite{matusik2000image,starck2007surface,waschbusch2005scalable}, or a temporally coherent shape by deforming a template to match the multi-view constraints~\cite{Carranza:2003,de2008performance}.
While some of the methods work for general scenes, many of them are dedicated to human bodies~\cite{gall2009motion,vlasic2008articulated,liu2011markerless,wu2012full,bray2006posecut,brox2010combined,wu2013onset,mustafa2015iccv}.
In recent years, many attempts have been made to alleviate the requirement of multi-camera systems by using depth sensors~\cite{Li2009,Zhang2014,Bogo_2015_ICCV,Guo_2015_ICCV,Helten:2013} or even a monocular RGB camera~\cite{MuVS:3DV:2017, Xu:2018:MHP:3191713.3181973, Habermann:2019:LRH:3313807.3311970, deepcap}.
Although compelling results have been demonstrated, these approaches are fundamentally ill-posed and suffer from occlusion and depth ambiguities.
Furthermore, in contrast to our method, they typically treat the character as a topologically connected template, and therefore are not able to handle movement of the clothing, such as sliding of the sleeves on the arms.
Another line of work specifically focuses on capturing clothing deformations~\cite{pons2017clothcap,bradley2008markerless,zhou2013garment,chen2015garment,xiang2020MonoClothCap}.
For instance, ClothCap~\cite{pons2017clothcap} automatically segments the different pieces of clothing and tracks the deformation of the clothing over time from 4D scans. Zhang and colleagues recover the detailed body shape under the clothing~\cite{zhang2017detailed}.
Our approach relies on these two methods for the generation of training data. More recently, multiple approaches have been proposed to capture human appearance by modeling the radiance field with a deep neural network~\cite{Peng_2021_CVPR,park2020deformable,pumarola2021d,wang2021learning}. These methods can synthesize photorealistic novel views of the captured scene or human subject, but unlike our work, cannot be used as animatable virtual avatars.

%% file: method.tex
\begin{figure}[t]
    \centering
    \includegraphics[width=\linewidth]{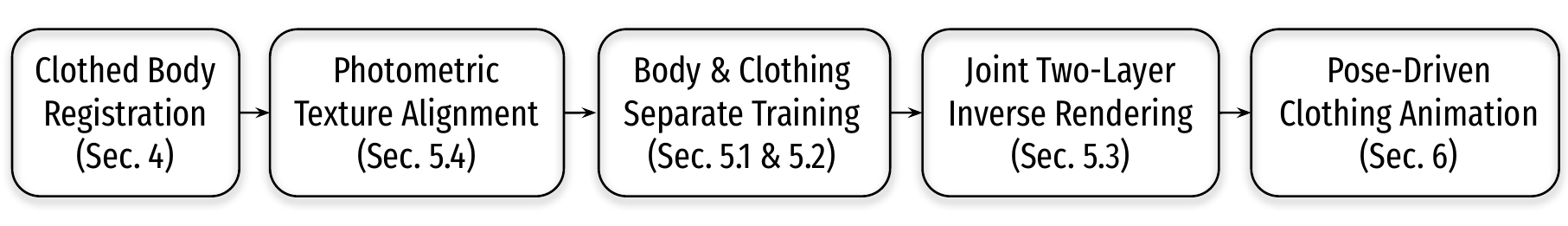}
    \caption{An overview of our proposed method in procedural order.}
    \label{fig:method-overview}
\end{figure}


\begin{figure*}[t]
    \centering
    \includegraphics[width=\linewidth]{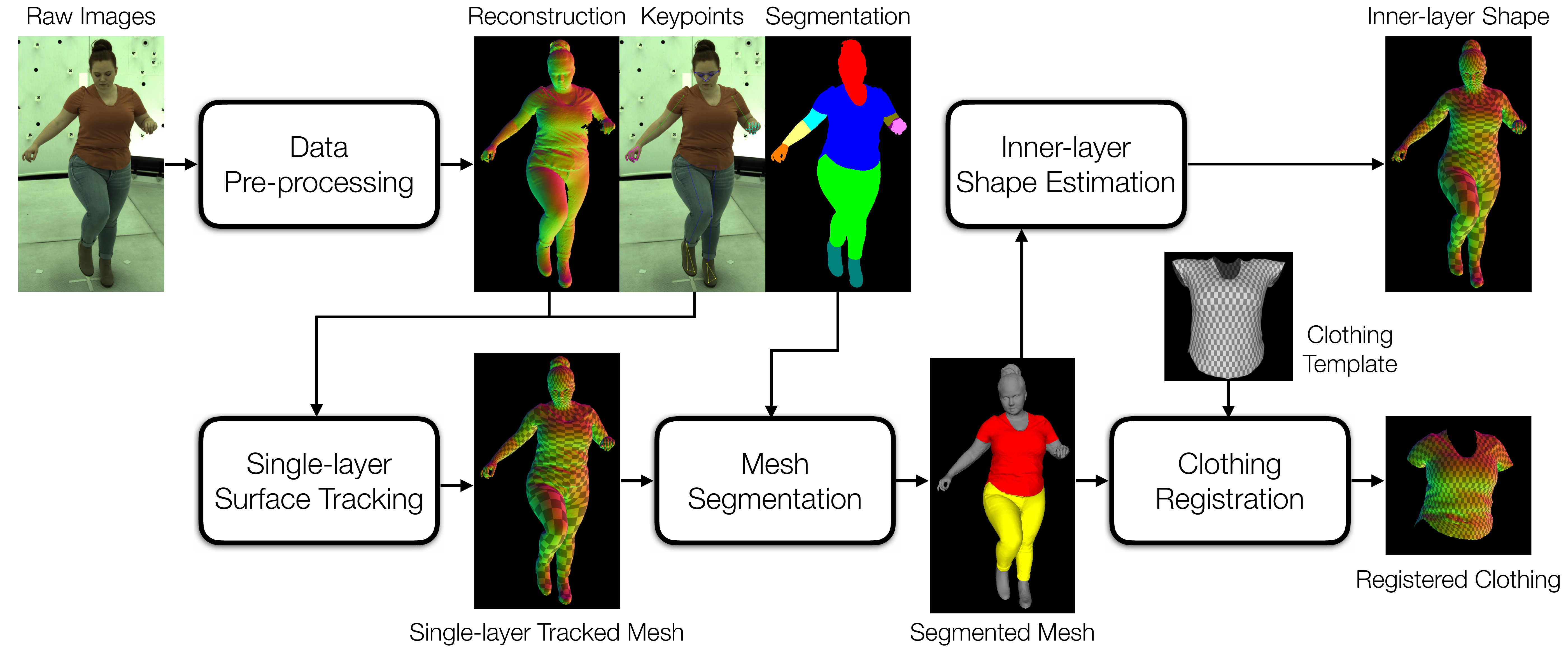}
    \caption{The clothed body registration pipeline that we use to generate training data for our two-layer codec avatars.}
    \label{fig:clothed-body-registration}
\end{figure*}

\section{Method Overview}

Our goal in this paper is to build full-body clothed digital avatars that enable photorealistic rendering from any viewpoint. To make the avatars useful, they should be animatable given some driving signals that can be obtained at modest cost. We choose 3D skeletal joint angles and facial keypoints as the input conditioning, similar to previous work \cite{bagautdinov2021driving}. For example, these driving signals can be obtained by multi-view triangulation and inverse kinematics from a sparse set of cameras.

The central idea of our method is to explicitly represent body and clothing as two separate layers. We take this approach for three reasons. First, we notice that the deformation of the body and the clothing follow different movement patterns because of their different dynamics. A single frame of joint angles in the driving signal can largely determine the body state through Linear Blending Skinning (LBS) and pose-dependent deformation. In contrast, the dynamics of clothing can vary too much to be described only by current body pose without considering temporal information. Thus the body and clothing layers need to be controlled by different input conditioning. Second, in the single-layer registration of the body with the clothing, a specific vertex along the clothing boundary can belong to either the body region or the clothing region in different frames due to the sliding motion of the clothing relative to the body, which violates the single layer assumption. A codec avatar trained with such data often has a color between the clothing and skin colors in such a region, leading to ghosting effects around the sleeves and neck of the garment. Although disentanglement could alleviate this kind of artifact, it cannot eliminate it due to limited training data capturing the complex interaction between clothing and the body. In our work, with the registration of body and clothing in separate layers, such artifacts can be avoided because each vertex is either part of body or the clothing across all frames. Third, separate layers for body and clothing open up opportunities for further changing the appearance of the avatar, such as temporally consistent editing of the clothing texture without interfering with the body appearance. This capability might also make it possible to alter the clothing style through physical simulation, which we leave for future work.

In this work, we assume that the subject to be modeled wears a T-shirt and pants. We only model the T-shirt in the second, outer layer because it exhibits most of the dynamics and variations in geometry and texture. In the inner layer, we model the body region covered by the outer layer (torso and upper arms) and the rest of human surface, including the head, arms, pants\footnote{The pants of the captured subjects in this work are tight and thus not worth the effort of modeling as a separate layer. We demonstrate in the results that the advantage of clothing modeling as a separate layer is obvious when the garment is loose.} and shoes.

In Section~\ref{sec:dataProcessing}, we briefly describe our two-layer geometry-based surface registration method to generate the necessary training data for the codec avatars. In Section~\ref{sec:codecAvatar}, we present our two-layer codec avatars. We describe the architecture of the body branch in Section~\ref{sec:bodydecoder} and clothing branch in Section~\ref{sec:clothdecoder}, as well as the joint training of both branches through inverse rendering in Section~\ref{sec:ir}. In Section~\ref{sec:textureAlignment}, we propose a method for texture alignment to improve the photometric correspondences between registered clothing meshes across different frames. In Section~\ref{sec:animation}, we present the temporal model used to animate our clothed avatars using a sequence of joint angles as the driving signal. A visualization of the method is shown in Fig.~\ref{fig:method-overview}.

\begin{figure*}[t]
    \centering
    \includegraphics[width=\linewidth]{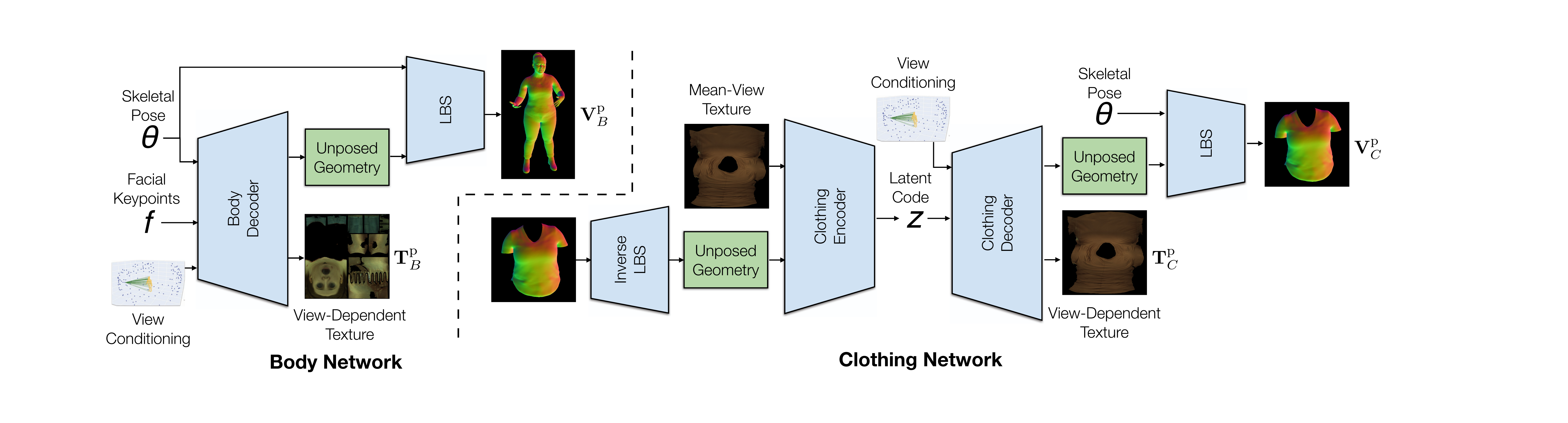}
    \caption{Network architecture of our two-layer full-body codec avatar. We show the body network on the left and the clothing network on the right, including the input and output of each network.}
    \label{fig:combined-network}
\end{figure*}

\section{Clothed Body Registration}
\label{sec:dataProcessing}

The pipeline to generate the data for training our two-layer codec avatars is illustrated in Fig.~\ref{fig:clothed-body-registration}. Our goal is to register the body and clothing geometry in two separate layers. A more detailed description of this pipeline can be found in the supplementary document.

\textit{Data preprocessing.} The input to our pipeline is a sequence of RGB images of the subject captured by a synchronized multi-camera system. The raw RGB images are used to create a dense 3D reconstruction of the human surface with a multi-view Patchmatch reconstruction algorithm~\cite{Galliani15}. An example of the reconstructed mesh can be seen in Fig.~\ref{fig:clothed-body-registration}. In addition, we obtain a part segmentation of different body and clothing regions for each captured image. We also run 2D keypoint detection for the body, face and hands, which are triangulated to obtain 3D keypoints.

\textit{Single-Layer Surface Tracking.} We non-rigidly register the reconstructed meshes with a kinematic body model, similar to \cite{zhang2017detailed} and \cite{walsman2017}. We estimate a personalized rest-state shape and a set of of joint angles for each frame by minimizing the difference between the LBS output and the reconstructed surface, as well as the 3D keypoints in the previous step. We further perform free-form Iterative Closest Points (ICP) registration using the skinned kinematic model as initialization. 

\textit{Mesh Segmentation.} In this step, we segment the single-layer tracked meshes into separate body and clothing parts. We unproject the image segmentation labels onto the mesh and for each vertex take the majority of votes across different views. Similar to \cite{pons2017clothcap}, we also run the Markov Random Field (MRF) to remove noisy segmentation labels.

\textit{Clothing Registration.} Our clothing registration step is similar to \cite{pons2017clothcap}. We manually create a template clothing mesh and use it to register the clothing region of the single-layer tracked mesh for each frame. Essentially we run a non-rigid ICP algorithm that aligns the template and target clothing region. To provide good initialization for the optimization, we find it useful to apply Biharmonic Feformation Fields~\cite{jacobson2010mixed} which generate a deformed template mesh whose boundary is directly aligned with the target clothing boundary with the lowest possible interior distortion.

\textit{Inner-Layer Shape Estimation.} The inner-layer geometry consists of two parts: the invisible body region covered by the clothing in the upper body, which we estimate using the method in \cite{zhang2017detailed}, and the visible region of the human surface, which can be directly obtained by matching with the single-layer tracking results. Unlike \cite{zhang2017detailed}, we only need to estimate the underlying body shape of the upper body, because the pants are treated as part of the inner layer in this work.

\section{Clothed Body Modeling}
\label{sec:codecAvatar}

We now present our two-layer codec avatars with explicit clothing modeling. Similar to \cite{lombardi2018deep} and \cite{bagautdinov2021driving}, we employ a Variational Autoencoder (VAE) as our generative model. In our two-layer formation, we train a separate network to learn the deformation space for body and clothing, while the correlation between body and clothing can be learned afterwards with a temporal model for animation. To this end, we train a body decoder which takes the skeletal pose as input, and predicts geometry and view-conditioned texture for the inner body layer, as shown on the left of Fig.~\ref{fig:combined-network}. Similarly, we train a clothing decoder with a VAE, as shown on the right of Fig.~\ref{fig:combined-network}. Similar to existing approaches to body modeling~\cite{loper2015smpl,STAR:2020}, we only learn the geometry in the canonical pose space for both the body layer and the clothing layer by applying an inverse LBS transform. This technique reduces the deformation space that needs to be learned. In the following sections, we introduce the detailed structure for the body and clothing networks, and explain how we train them.
Implementation details including loss weights and network architecture can be found in the supplementary document.

\subsection{Body Decoder}
\label{sec:bodydecoder}

As shown on the left of Fig.~\ref{fig:combined-network}, our body network is similar to the decoder structure in \cite{bagautdinov2021driving}, without the encoder. Once the clothing is decoupled from the body, the skeletal pose and facial keypoints contain sufficient information to describe the body state (including pants that are relatively tight). We do not use a latent code as conditioning for the body network to avoid the difficult problem of disentanglement between the latent space and the driving signal, as described in \cite{bagautdinov2021driving}. Our body decoder takes in the skeletal pose, facial keypoints and view-conditioning as input, produces unposed geometry in a UV positional map and view-dependent texture for the body as output. A LBS transformation is then applied to the unposed mesh restored from the UV map to produce the final output mesh.


The loss function to train the body network is defined as:
\begin{equation}
\begin{aligned}
    E^B_\text{train} ={} &\lambda_g \Vert \mathbf V_B^\text{p} - \mathbf V_B^\text{r} \Vert^2 + \lambda_{lap} \Vert \text{L}(\mathbf V_B^\text{p}) - \text{L}(\mathbf V_B^\text{r}) \Vert^2 \\
    & + \lambda_t \Vert (\mathbf T_B^\text{p} - \mathbf T_B^\text{t}) \odot M_B^\text{V} \Vert^2,
    \label{eq:bodyloss}
\end{aligned}
\end{equation}
where $\mathbf V_B^\text{p}$ is the vertex position interpolated from the predicted position map in UV, and $\mathbf V_B^\text{r}$ is the vertex from inner layer registration from Sec.~\ref{sec:dataProcessing}, $L(\cdot)$ is the Laplacian operator, $\mathbf T_B^\text{p}$ is the predicted texture, $\mathbf T_B^\text{t}$ is the reconstructed texture per-view, and $M_B^\text{V}$ is the mask indicating the valid UV region.



\subsection{Clothing Network}
\label{sec:clothdecoder}

As shown on the right of Fig.~\ref{fig:combined-network}, we model the clothing appearance with a Conditional Variational Autoencoder (cVAE). The encoder takes as input the unposed clothing geometry and mean-view texture, and produces parameters of a Gaussian distribution, from which a latent code $\mathbf z$ is sampled. Besides the latent code, the decoder also takes spatial-varying view conditioning as input, and predicts geometry and texture for the clothing. Then, the training loss is described as: 
\begin{equation}
\begin{aligned}
    E^C_\text{train} ={} & \lambda_g \Vert \mathbf V_C^\text{p} - \mathbf V_C^\text{r} \Vert^2 +\lambda_{lap} \Vert \text{L}(\mathbf V_C^\text{p}) - \text{L}(\mathbf V_C^\text{r}) \Vert^2 \\
    & + \lambda_t \Vert (\mathbf T_C^\text{p} - \mathbf T_C^\text{t}) \odot M_C^\text{V} \Vert^2 + \lambda_{\text{kl}} E_{\text{kl}},
    \label{eq:clothloss}
\end{aligned}
\end{equation}
where $\mathbf V_C^\text{p}$, $\mathbf V_C^\text{t}$, $\mathbf T_B^\text{p}$, $\mathbf T_B^\text{t}$, and $M_C^\text{V}$ are all defined similarly to the parameters in the body decoder but with respect to clothing, $E_{\text{kl}}$ is a conventional KL divergence loss.

\subsection{Inverse Rendering with Two-layer Representation}
\label{sec:ir}

The ICP-based clothing registration algorithm in Section \ref{sec:dataProcessing} and previous work \cite{pons2017clothcap} aims to align the boundary of the clothing template with the target area, while there is no explicit constraint for the interior correspondences except for the mesh regularization. Therefore, the registered meshes from Sec.~\ref{sec:dataProcessing} may suffer from correspondence errors in the interior (see the first column of Fig. \ref{fig:ir-results}), which significantly influences the decoder quality, especially for dynamic clothing. In order to correct the correspondences in the training stage, we need to link the predicted geometry and texture to the input multi-view images in a differentiable way.
To this end, after the body and clothing networks are separately trained as described in Sec.~\ref{sec:bodydecoder} and \ref{sec:clothdecoder}, we jointly train the body and clothing networks by rendering the output with a differentiable renderer. We use the following loss functions:
\begin{equation}
\begin{aligned}
    E^{\text{inv}}_\text{train} ={} &\lambda_i \Vert \mathbf I^\text{R} - \mathbf I^\text{C} \Vert + \lambda_m \Vert \mathbf M^\text{R} - \mathbf M^\text{C} \Vert\\
    &+ \lambda_{v} E_{\text{softvisi}} + \lambda_{lap} E_{\text{lap}},
\end{aligned}
\end{equation}
where $\mathbf I^\text{R}$ and $\mathbf I^\text{C}$ are the rendered image and the captured image, $\mathbf M^\text{R}$ and $\mathbf M^\text{C}$ are the rendered foreground mask and the captured foreground mask, and $E_{\text{lap}}$ is the Laplacian geometry loss similar to that defined in Eqn.~\ref{eq:bodyloss} and \ref{eq:clothloss}. $E_{\text{softvisi}}$ is a soft visibility loss, similar to \cite{softrasterizer19}, that is specifically designed to handle the depth reasoning between the body and clothing so that the gradient can be back-propagated through if the depth order is wrong. In detail, we define the soft visibility for a specific pixel as
\begin{gather}
    S = \sigma\left(\frac{D^\text{C} - D^\text{B}}{c}\right),
\end{gather} where $\sigma(\cdot)$ is the sigmoid function, $D^\text{C}$ and $ D^\text{B}$ are the depth rendered from the current viewpoint for the clothing and body layer, and $c$ is a scaling constant. Then the soft visibility loss is defined as:
\begin{gather}
    E_{\text{softvisi}} = S^2,
\end{gather}
when $S>0.5$ and the current pixel is assigned to be clothing according to the 2D cloth segmentation. Otherwise, $E_{\text{softvisi}}$ is set to $0$. If the pixel is labeled as clothing but the body layer is on top of the clothing layer from this viewpoint, the soft visibility loss will back-propagate the information to update the surfaces until the correct depth order is achieved.

Following \cite{bagautdinov2021driving} in this inverse rendering stage, we also use a shadow network that computes quasi-shadow maps for body and clothing given the ambient occlusion maps. In contrast to the approach of \cite{bagautdinov2021driving} where the ambient occlusion is approximated with the body template after the LBS transformation, we compute the exact ambient occlusion using the output geometry from the body and clothing decoders because we aim to model a more detailed clothing deformation than can be produced by the LBS transformation. The quasi-shadow map is then multiplied with the view-dependent texture before applying the differentiable renderer.

\begin{figure}[t]
    \centering
    \includegraphics[width=\linewidth]{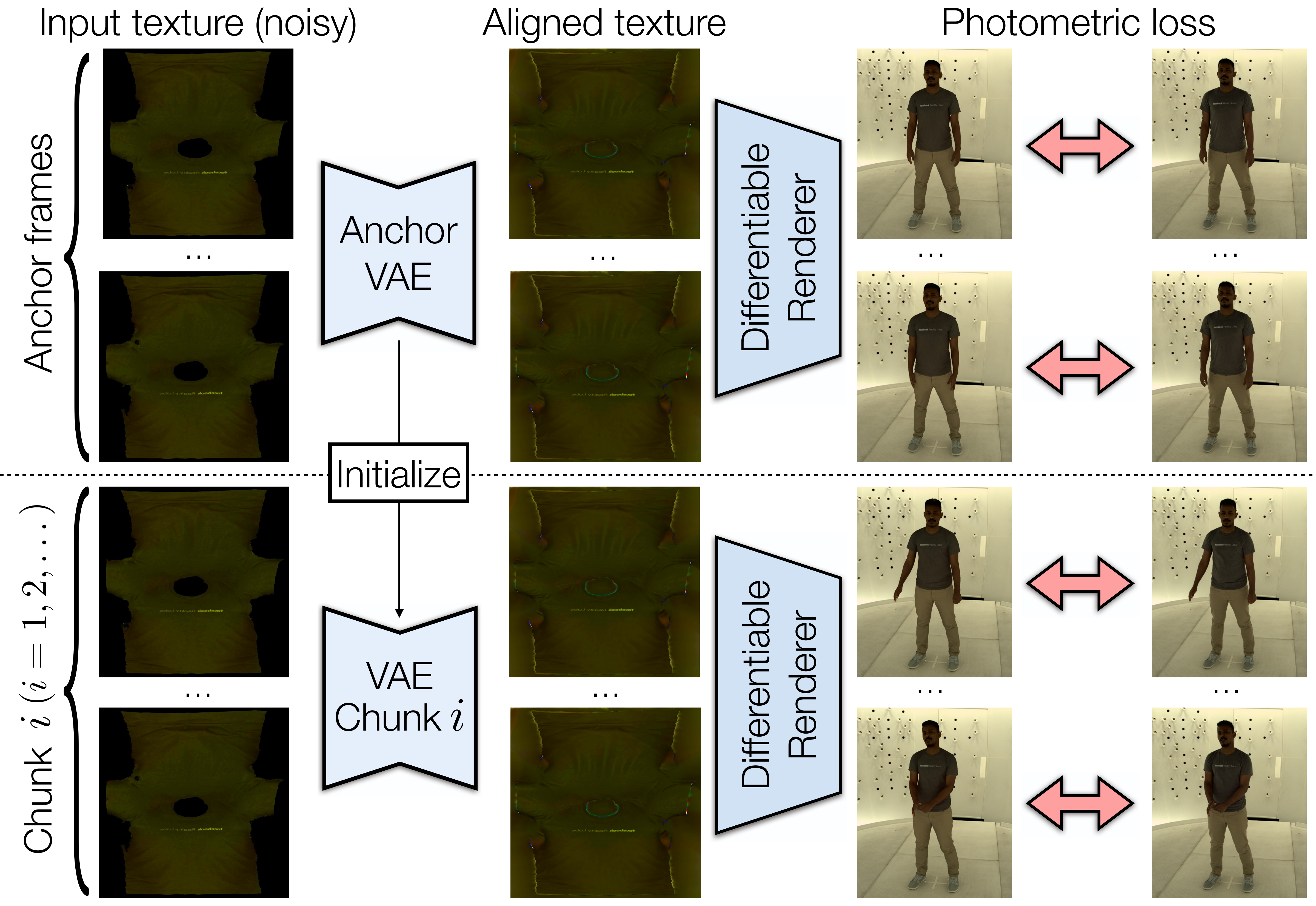}
    \caption{Our inverse-rendering-based photometric texture alignment method (Sec.~\ref{sec:textureAlignment}). First, the anchor frames are used to train the anchor VAE with photometric loss applied to the differentiable rendering output. Then, a separate VAE for each chunk of frames is initialized independently from the anchor VAE and trained using the same loss function. Here we only show the texture and omit the geometry in the VAE input and output for clarity.}
    \label{fig:texture-alignment}
\end{figure}

\subsection{Texture Alignment with Inverse Rendering}
\label{sec:textureAlignment}


The inverse rendering method mentioned in Sec.~\ref{sec:ir} already has the capability to improve photometric correspondences to some extent, because the network tends to predict texture with less variance across frames, along with deformed geometry to align the rendering output with the ground truth images. Ideally we only need to train the two decoders simultaneously with the inverse rendering loss to correct the correspondences while creating the generative model for driving the animation. However, we find that this alone would not correct all the correspondence errors. The model might not find a good minimum for two reasons. First, the variation in photometric correspondences in our initial registration may be too large for the network to fix. Secondly, our VAE model with view conditioning may allow the decoder to cheat with the view-dependent texture rather than moving the geometry.

\begin{figure*}[t]
    \centering
    \includegraphics[width=\linewidth]{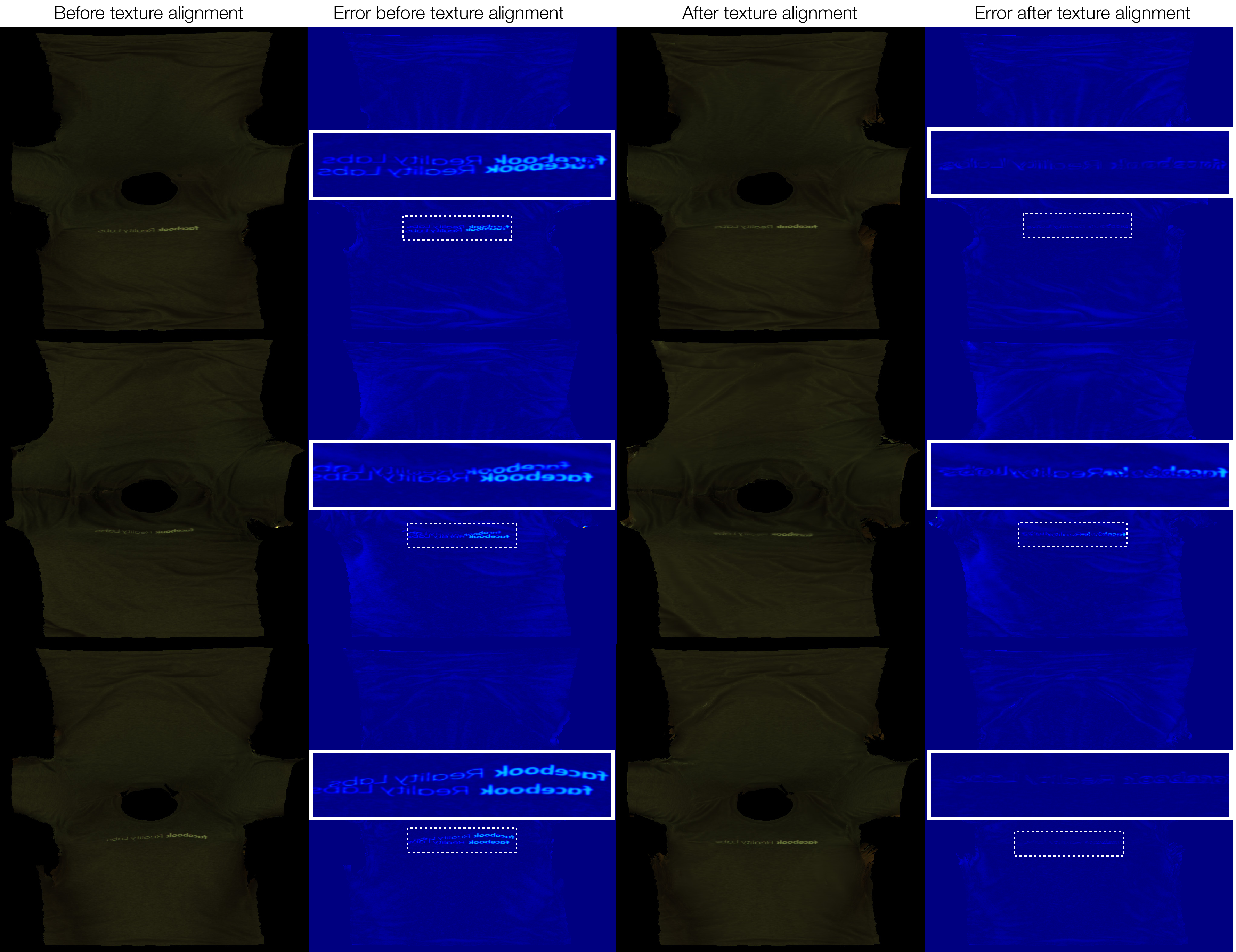}
    \caption{Inverse-rendering-based texture alignment results. From left to right, we show (1)~projected texture on the clothing mesh before texture alignment, (2)~error map between the first column and the mean texture of anchor frames, (3)~projected clothing texture after texture alignment, and (4)~the difference between the third column and the mean texture of anchor frames. The error maps are visualized with the Jet colormap; lighter color represents larger error. We also show a zoomed-in version of the text region to highlight the difference.}
    \label{fig:ir-results}
\end{figure*}

These problems motivate us to propose a new way to use inverse rendering for correspondence improvement. First, we separate the registered meshes into chunks of $50$ neighboring frames. Then, we select the first chunk as the anchor frames, and train an anchor network for this chunk using the inverse rendering model described in Sec.~\ref{sec:ir}. After convergence, we use the trained network parameters to initialize the training of other chunks. To make sure that the alignment of the other chunks does not drift from the anchor frames, we set a small learning rate (1e-4 for the AdamW optimizer), and mix the anchor frames with each other chunk during training. We remove the view conditioning from the texture branch of our decoder in Sec.~\ref{sec:ir}, and use a single texture prediction for inverse rendering in all the camera views. The output geometry predicted by the network of each chunk after training has more consistent correspondences across frames compared with the input, which is manifested by the consistent projected texture pattern in the UV space shown in Fig.~\ref{fig:ir-results}. A visual illustration of this process is provided in Fig.~\ref{fig:texture-alignment}. This method has a similar spirit to previous UV-template-based texture alignment approaches \cite{garrido2013reconstructing,Bogo_2017_CVPR}, but naturally extends the idea to a neural-network formulation under the framework of codec avatars.

The method described here is applied after the two-layer registration is obtained in Section~\ref{sec:dataProcessing}, as shown in Fig.~\ref{fig:method-overview}. For each frame, we use the output geometry predicted by the network as a new registered mesh with the improved correspondences. We use these data to train the body and the clothing networks, as described in Section~\ref{sec:bodydecoder}-\ref{sec:ir}.  

\begin{figure}[t]
    \centering
    \includegraphics[width=\linewidth]{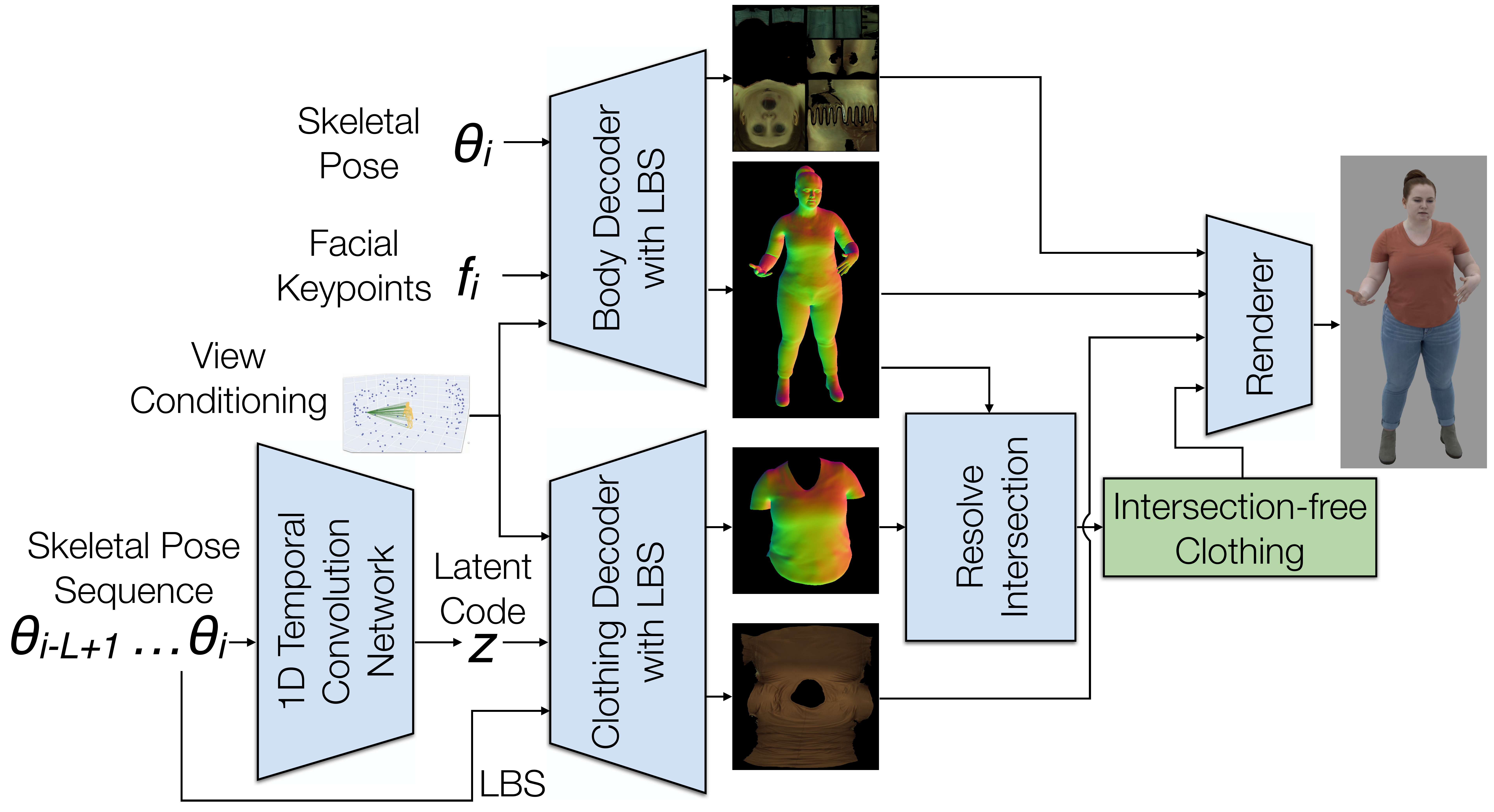}
    \caption{The clothed body animation pipeline.}
    \label{fig:animation-pipeline}
\end{figure}

\begin{figure}[t]
    \centering
    \includegraphics[width=\linewidth]{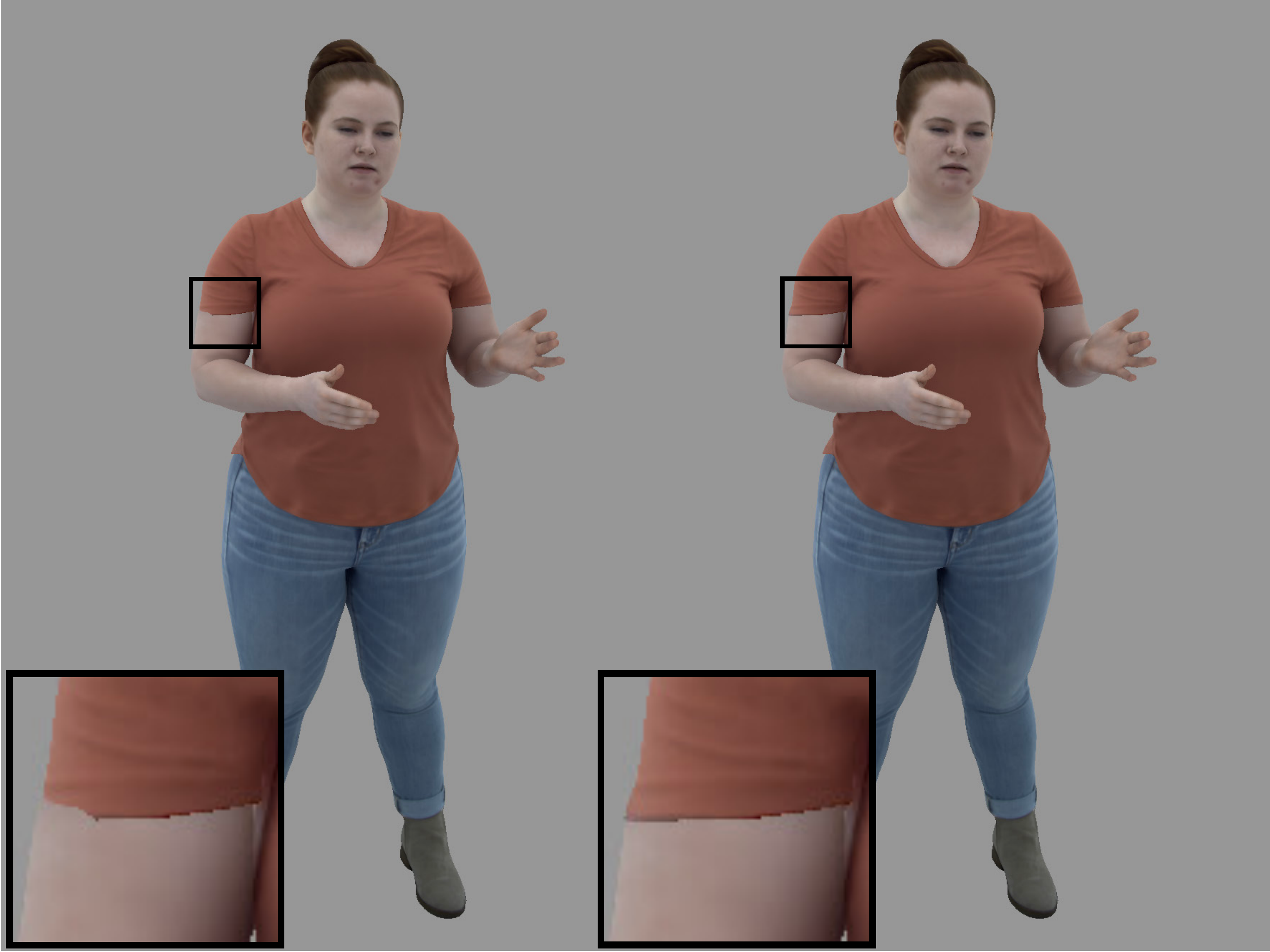}
    \caption{An example of resolving intersection. The intersecting area is highlighted by the zoomed boxes.}
    \label{fig:collision}
\end{figure}

\section{Temporal Modeling for Pose-Driven Clothing Animation}
\label{sec:animation}

In our two-layer codec avatars, the body output is conditioned on a single frame of skeletal pose and facial keypoints, while the clothing state is determined by the latent code. In order to animate the clothing from the driving signal, we use a Temporal Convolution Network (TCN) to learn the correlation between body dynamics and clothing deformation. Our TCN takes in the sequence of previous and current skeletal pose and infers the latent clothing state.

An illustration of our animation pipeline is shown in Fig.~\ref{fig:animation-pipeline}. The temporal convolution network takes as input the joint angles in a window of $L$ frames up to the target frame, and passes through several 1D temporal convolution layers to predict the clothing latent code for the current frame $\mathbf z$. To train the TCN, we minimize the following loss function:
\begin{gather}
    E^{TCN}_{\text{train}} = \Vert \mathbf z - \mathbf z^\text{c} \Vert^2,
\end{gather}
where $\mathbf z^\text{c}$ is the ground truth latent code obtained from the trained clothing VAE.

An alternative formulation would be to condition the prediction on not just previous body states, but also previous clothing states. This formulation is inspired by cloth simulation, where the clothing vertex position and velocity in the previous frame are needed to compute the current clothing state. However, in our data-driven setting, we find that such an auto-regressive model that takes in previous clothing states is hard to train and does not outperform the non-autoregressive model given the limited amount of data (25 min). Therefore, the input to our TCN is a temporal window of skeletal poses, not including the previous clothing states.

\paragraph{Resolving Intersection.} One solution is to add a training loss for TCN to make sure that the predicted clothing does not intersect with the body. However, even without a loss to penalize intersection, the clothing states predicted by our TCN model already match the body shape well, resulting in only minimal intersection. Thus we only need to resolve intersection as a post processing step. We project the intersecting clothing back onto the body surface with an additional margin in the body normal direction. This operation will solve most intersections and make sure that the clothing and body are in the right depth order for rendering. An example of these results can be seen in Fig.~\ref{fig:collision}.


%% file: results.tex
\begin{figure}[t]
    \centering
    \includegraphics[width=\linewidth]{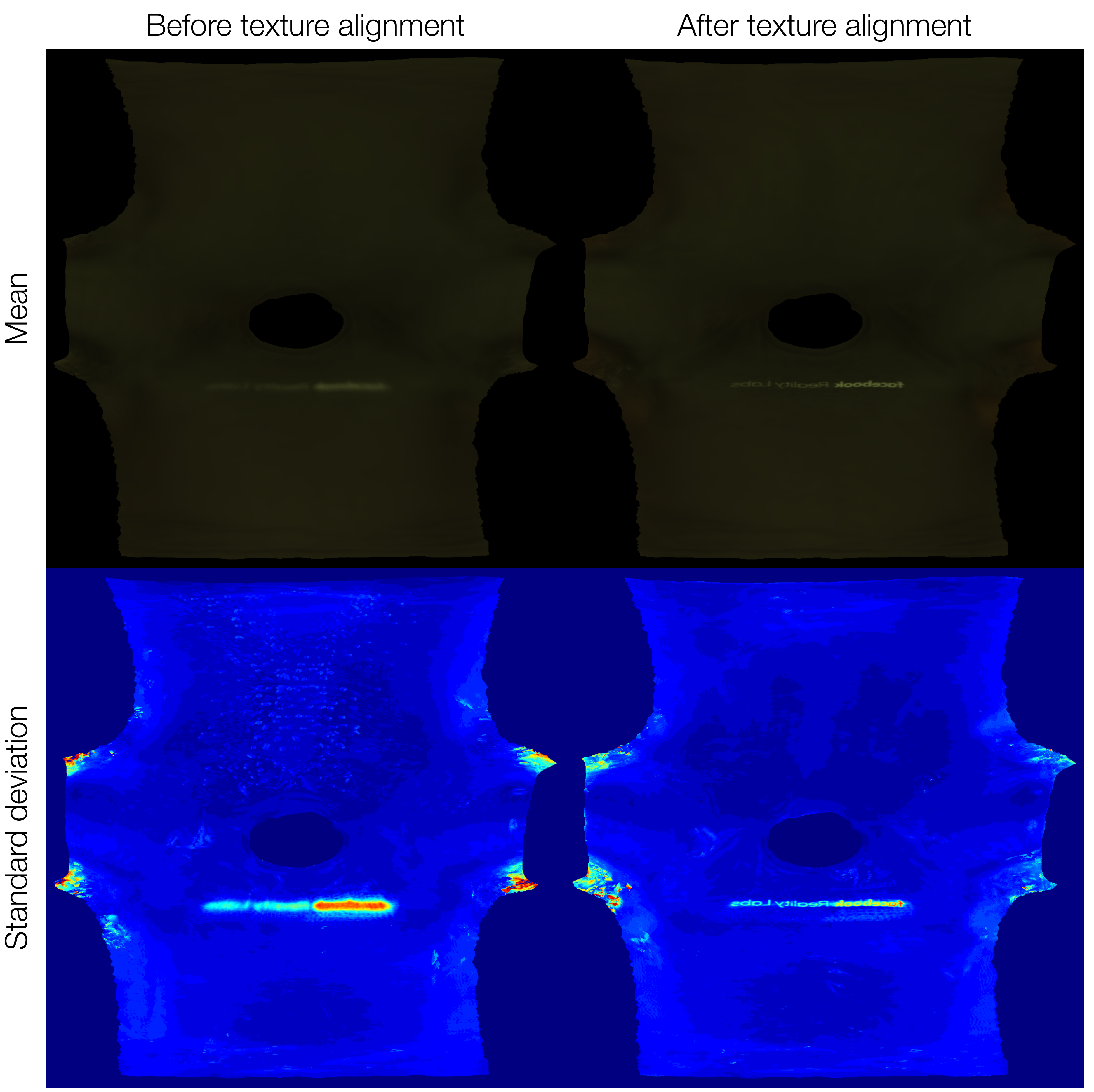}
    \caption{Mean (top row) and standard deviation (bottom row, converted to jet colormap) of unwrapped texture before (left column) and after (right column) texture alignment on the sequence of \textit{Subject 2}.}
    \label{fig:texture-mean-std}
\end{figure}

\begin{figure}[t]
    \centering
    \includegraphics[width=\linewidth]{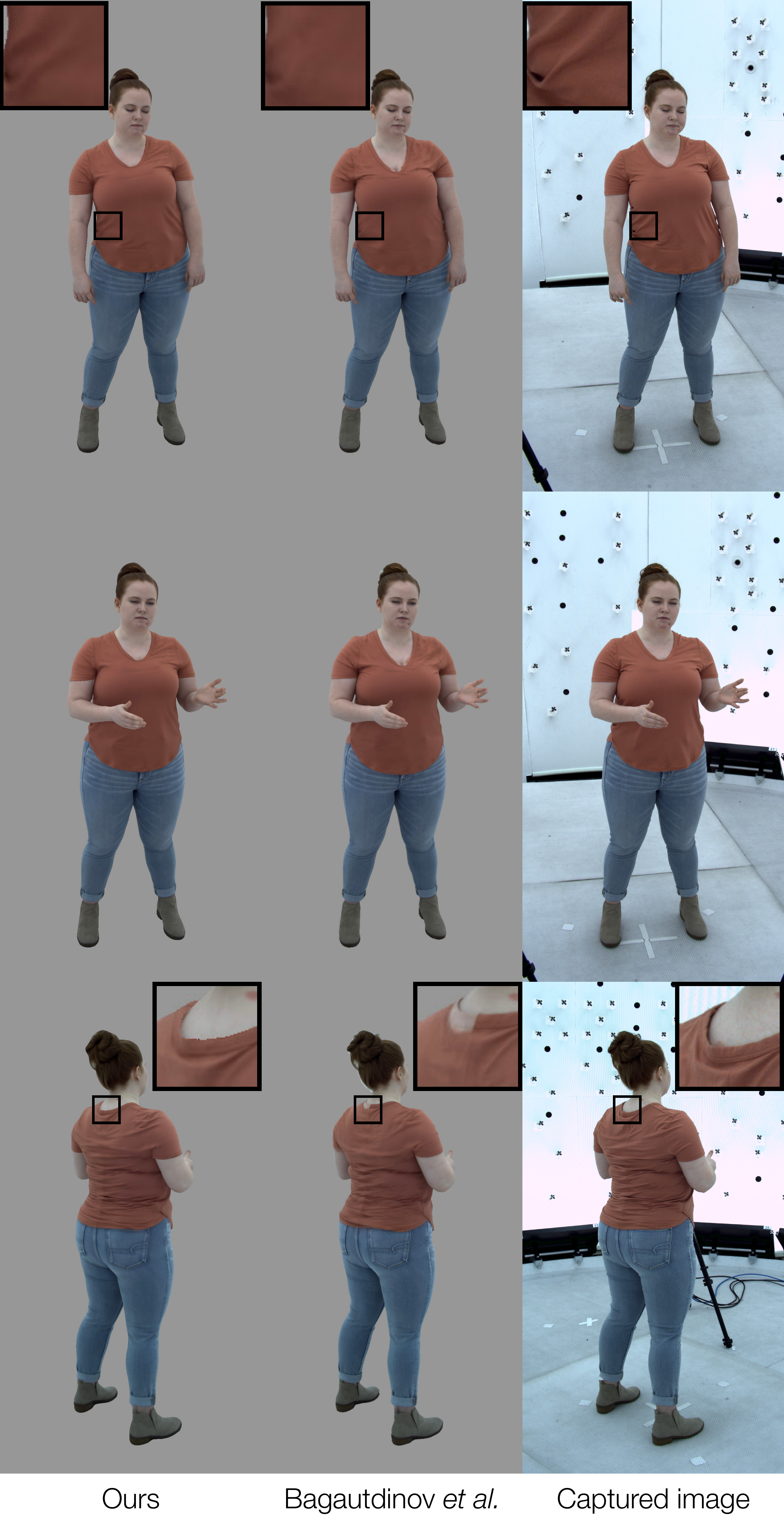}
    \caption{Comparison of animation output between our proposed method and baseline \cite{bagautdinov2021driving} on the \textit{Subject 1} sequence.}
    \label{fig:autumn-driving}
\end{figure}

\begin{figure}[t]
    \centering
    \includegraphics[width=\linewidth]{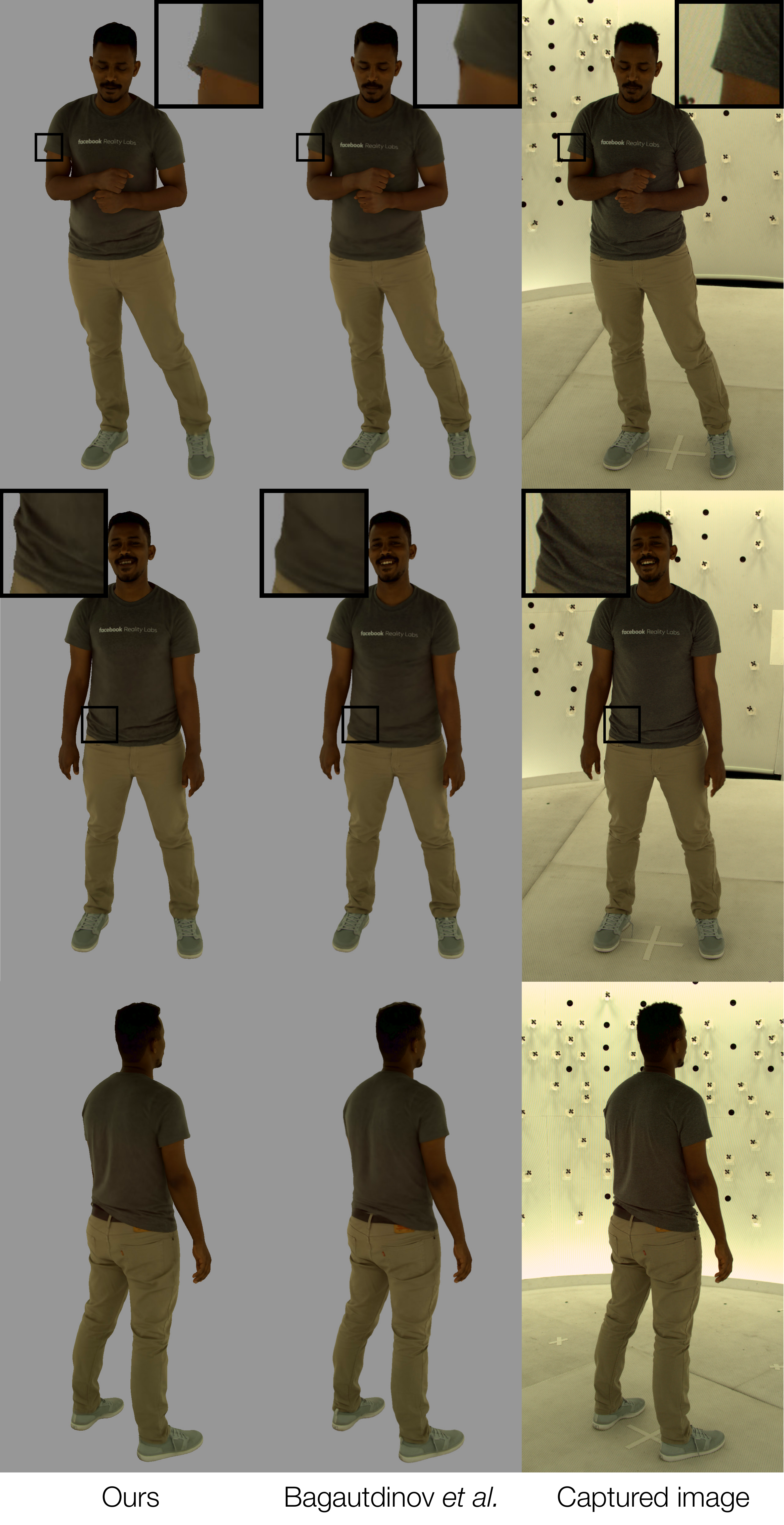}
    \caption{Comparison of animation output between our proposed method and baseline \cite{bagautdinov2021driving} on the \textit{Subject 2} sequence.}
    \label{fig:israel-driving}
\end{figure}

\begin{figure}[t]
    \centering
    \includegraphics[width=\linewidth]{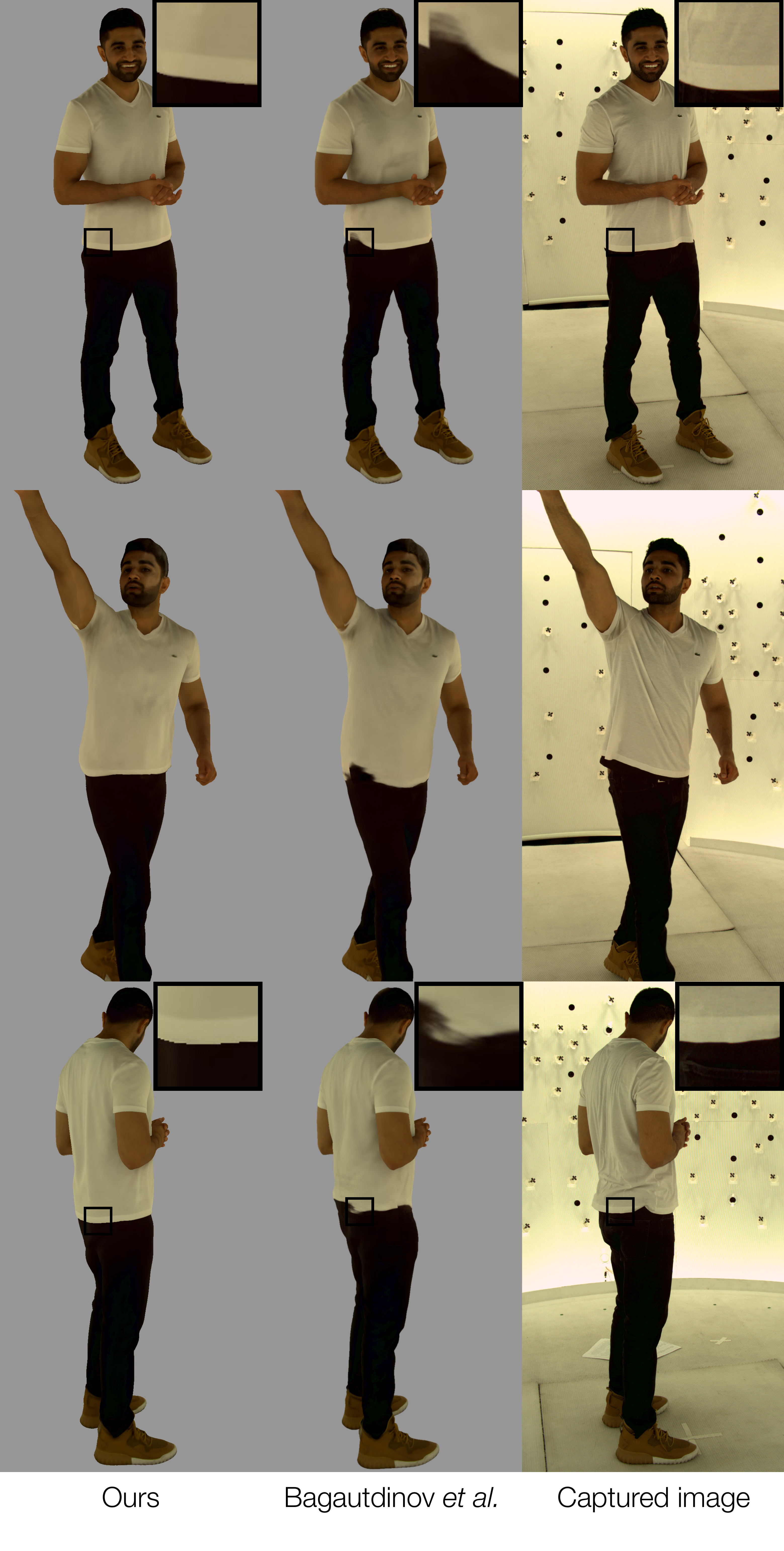}
    \caption{Comparison of animation output between our proposed method and baseline \cite{bagautdinov2021driving} on the \textit{Subject 3} sequence.}
    \label{fig:third-driving}
\end{figure}

\begin{figure*}[t]
    \centering
    \includegraphics[width=1.0\linewidth]{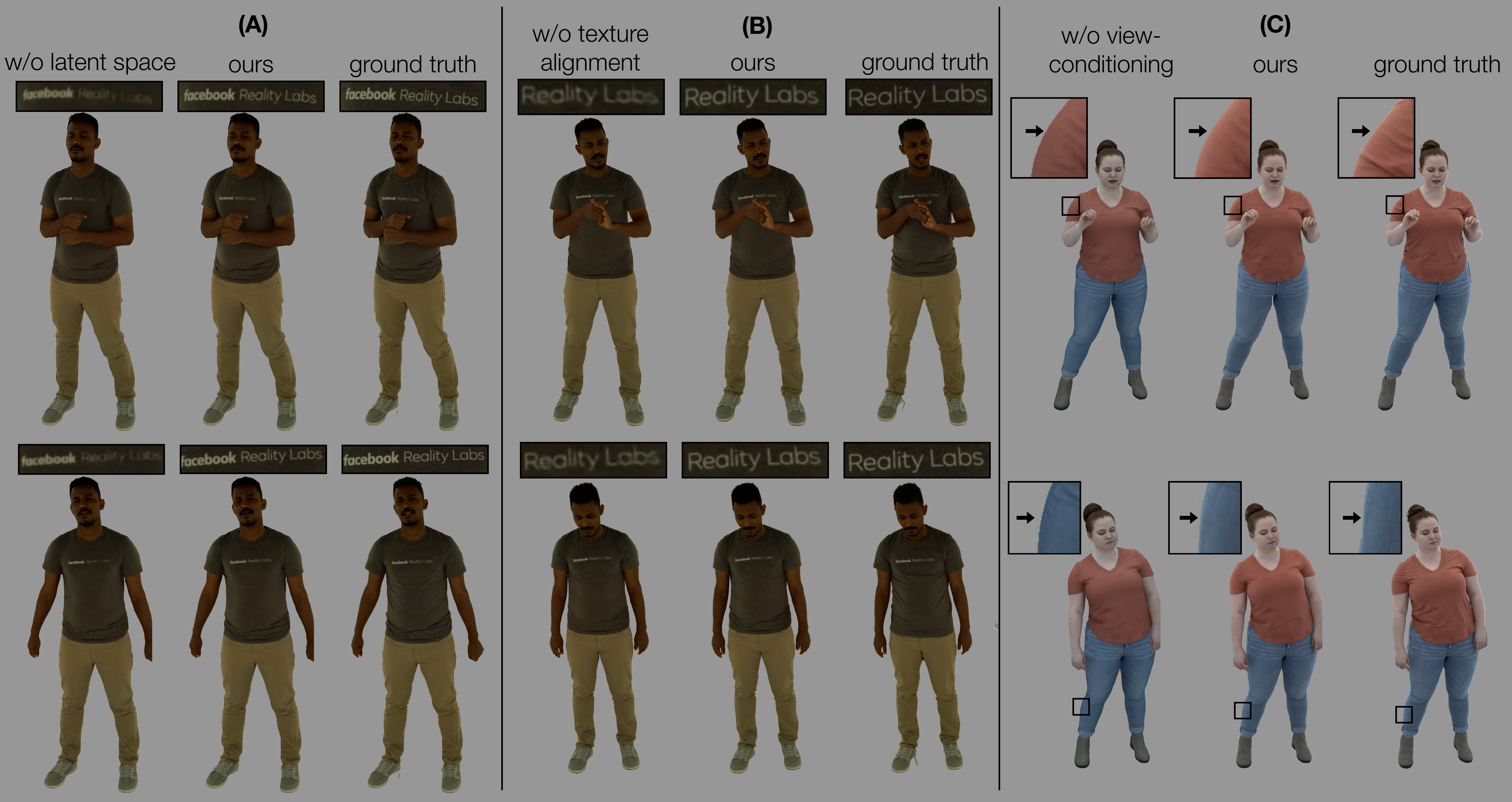}
    \caption{Ablation analysis of system components. In (A) we compare our results with a model without clothing VAE latent space for clothing, instead directly regressing clothing geometry and texture from a sequence of skeleton poses as input. In (B) our output is compared with the model trained using data without the texture alignment step. In both (A) and (B) our method shows sharper logo pattern. In (C), we show results with (ours) and without view-conditioning effects. Notice the strong reflectance of lighting near the silhouette of subject captured by our view-conditioning modeling.}
    \label{fig:ablation}
\end{figure*}

\begin{figure*}[t]
    \centering
    \includegraphics[width=0.9\linewidth]{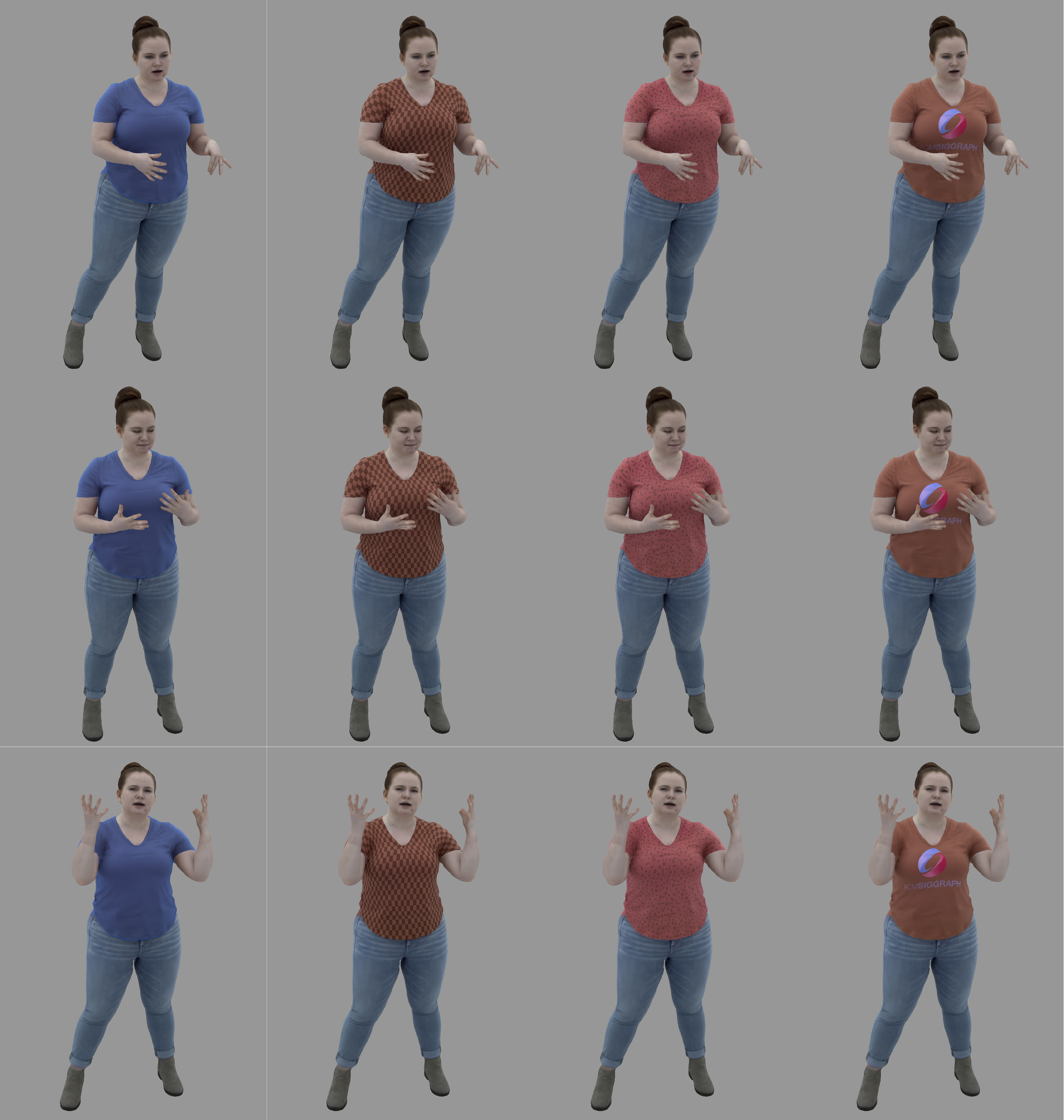}
    \caption{Texture editing results of our two-layer codec avatars. From left to right, we show application of color transformation, checkerboard pattern, random artist-created pattern, and an ACM SIGGRAPH logo, respectively, for three different frames.}
    \label{fig:texture-editing}
\end{figure*}

\begin{figure}[t]
    \centering
    \includegraphics[width=\linewidth]{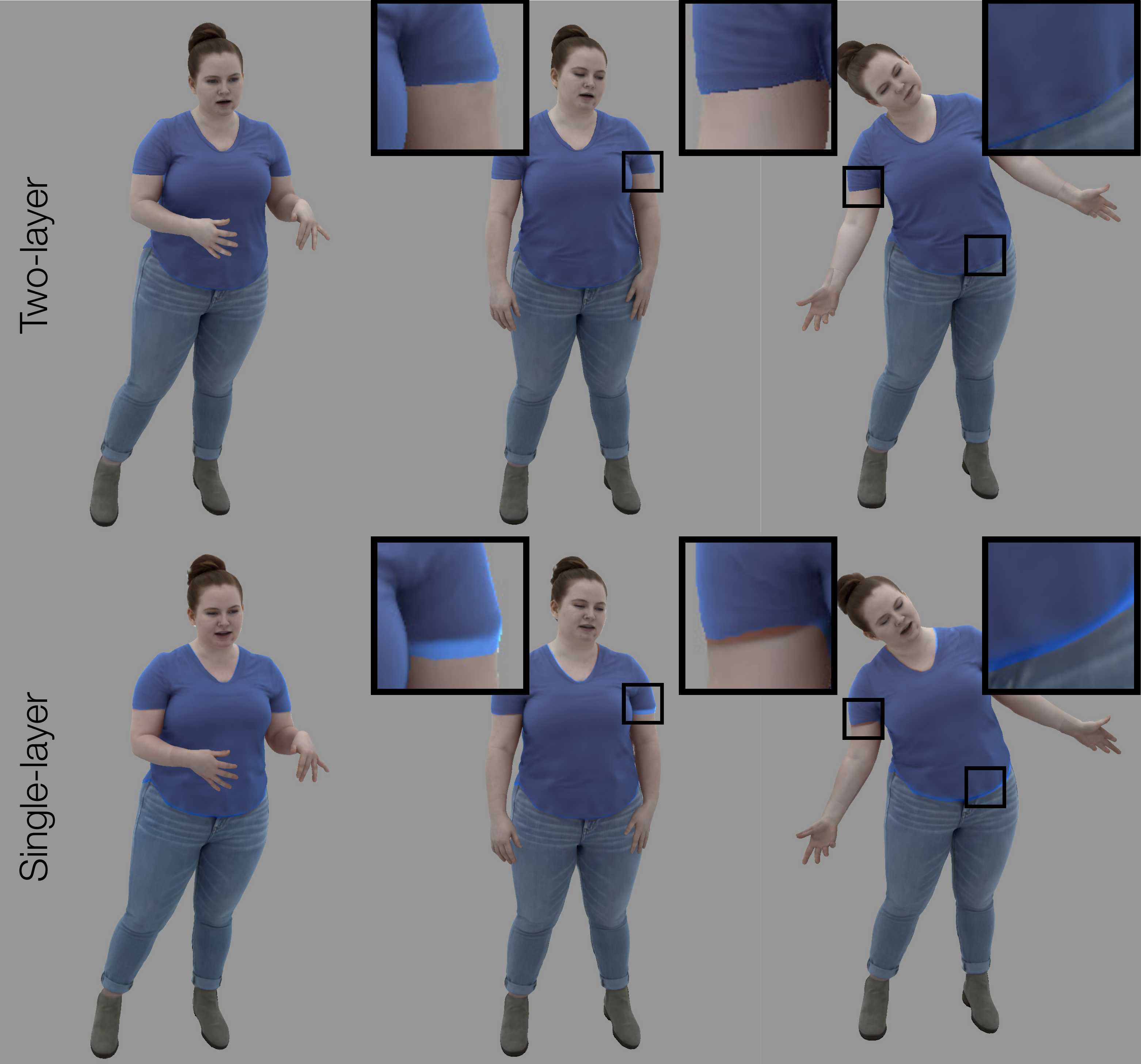}
    \caption{Comparison between the single-layer model (bottom row) and the two-layer model (top row) on texture editing in three different frames. The first column shows the frame where we manually segment out the upper clothing region in the UV space for the single-layer model.}
    \label{fig:texture-editing-compare-single}
\end{figure}


\section{Results}

In this section, we first introduce our capture system and captured data. Then we show the results of our photometric texture alignment method to demonstrate its effectiveness in achieving better photometric correspondence in the UV space. After that, we show the animation output of our two-layer codec avatars with explicit clothing modeling. In particular, we demonstrate the advantage of our two-layer formulation against the single-layer model in previous work. We close by demonstrating clothing texture editing for animation.

\subsection{Data Capture}

The training data for our codec avatars are captured by a multi-view capture system consisting of around 140 cameras that are distributed uniformly on a half dome above the ground. All the cameras run with hardware synchronization, capturing at the resolution of $4096 \times 2668$ and 30 fps. Three identities, one female (\textit{Subject 1}) and two males (\textit{Subject 2} and \textit{Subject 3}), are captured with a pre-defined acting script. The script is designed to capture peak poses with the activation going through all body joints, followed by a $10$-minute conversation to capture social behavior. For each subject, we collect sequences of 40k-50k frames in total and intentionally leave out approximately 4-5k contiguous frames for testing. 


\subsection{Texture Alignment with Inverse Rendering}
In this section, we show the results of texture alignment based on inverse rendering (Section \ref{sec:textureAlignment}) on the sequence of \textit{Subject 2}. Textures are projected from the raw captured images to the registered meshes before and after the texture alignment procedure, and then unwrapped into the UV space for comparison. Example results for several frames are shown in the first and third column of Fig.~\ref{fig:ir-results}. To assess the quality of alignment, we compare the mean UV texture of the anchor frames with the unwrapped texture of each individual frame. The error map is then visualized by the Jet colormap, shown in the second and fourth column of Fig.~\ref{fig:ir-results} respectively.

The visible pattern in the heatmap before texture alignment (the second column) verifies the lack of accurate interior correspondences in the registered clothing meshes from the ICP algorithm (Section \ref{sec:dataProcessing}). After the texture alignment (the fourth column), the error between the UV texture of those frames and the mean of anchor frames is significantly reduced. This result suggests that the correspondences in the mesh interior are improved in the inverse rendering process, and demonstrates the effectiveness of our texture alignment method. 

To statistically evaluate the quality of photometric correspondence in the UV space, we compute the mean and standard deviation of the unwrapped texture across different frames, as visualized in Fig.~\ref{fig:texture-mean-std}. Comparing the mean texture images, we observe a much sharper text pattern after texture alignment than before. Similarly, the standard deviation after texture alignment becomes smaller and more concentrated in the spatial domain. This result also verifies the improvement of photometric correspondence thanks to our texture alignment approach.

\subsection{Pose-Driven Animation}
\label{sec:eval-animation}

In this section, we present animation results produced by our two-layer codec avatars driven by the 3D skeletal pose and facial keypoints. In our animation pipeline, the body decoder is directly driven by skeletal pose and facial keypoints of the current frame; on the other hand, the clothing decoder is driven by latent clothing code generated by the temporal clothing model in Section~\ref{sec:animation}, which takes a temporal window of history and current poses as input. We compare the quality of our animation with previous work \cite{bagautdinov2021driving} that uses a single-layer codec avatar. We follow the method described in \cite{bagautdinov2021driving} to animate the single-layer codec avatar: we randomly sample the unit Gaussian distribution, and use the resulting noise values for imputation of the latent code. The sampled latent code, the skeletal pose and facial keypoints are fed together into decoder network. We present qualitative animation results on all three testing sequences, shown in Fig.~\ref{fig:autumn-driving}, \ref{fig:israel-driving}, and \ref{fig:third-driving}. Our animation results are better seen in the supplementary video.

Our two-layer formulation helps remove the severe artifacts in the clothing regions in the animation output of \cite{bagautdinov2021driving}, especially around the clothing boundary of Fig.~\ref{fig:autumn-driving}, and \ref{fig:third-driving}. Indeed, as the body and clothing are modeled together, the single-layer avatars rely on the latent code to describe the many possible clothing states corresponding to the same body pose. During animation, however, the absence of a ground truth latent code leads to degradation of the output, despite the efforts in \cite{bagautdinov2021driving} to disentangle the latent space from the driving signal. In contrast, our animation model achieves better animation quality by separating body and clothing into different modules: the body decoder can determine the body states given the driving signal of the current frame; the temporal model learns to infer the most plausible clothing states from body dynamics for a longer period; the clothing VAE ensures a reasonable clothing output given its learned smooth latent manifold. In addition, our two-layer avatars show results with a sharper clothing boundary and clearer wrinkle patterns in these images.

\begin{table}[t]
\centering
\begin{tabular}{c | C{1.3cm} C{1.3cm} | C{1.3cm} C{1.3cm}}
    \hline 
    \multirow{2}{*}{Sequence} & \multicolumn{2}{c|}{\cite{bagautdinov2021driving}} & \multicolumn{2}{c}{Ours}  \\
    \cline{2-5}
    \rule{0pt}{2.4ex} {} & MSE$\downarrow$ & SSIM$\uparrow$ & MSE$\downarrow$ & SSIM$\uparrow$  \\
    \hline
    \textit{Subject 1} & 100.57 & 0.8720 & \textbf{74.73} & \textbf{0.8816} \\
    \textit{Subject 2} & 81.95 & 0.8804 & \textbf{58.14} & \textbf{0.8917} \\
    \textit{Subject 3} & 456.20 & 0.8159 & \textbf{356.52} & \textbf{0.8230} \\
    \hline
\end{tabular}
\caption{Quantitative comparison between our proposed method and the previous work. We report Mean Square Error (lower better) and the Structural Similarity Index Measure (higher better) on all three testing sequences.}
\label{table:quantitative}
\end{table}

We also quantitatively compare the animation output of our two-layer codec avatars with the baseline method \cite{bagautdinov2021driving} by evaluating the output images against the captured ground truth images. We report the evaluation metrics of Mean Square Error (MSE) and Structural Similarity Index Measure (SSIM) over the foreground pixels. The results are shown in Tab.~\ref{table:quantitative}. Our method consistently outperforms \cite{bagautdinov2021driving} on all three sequences and both evaluation metrics. In particular, it is worth noting that our advantage on MSE is most obvious on the sequence of \textit{Subject 3}, who is wearing a loose T-shirt that is hard to model with the single-layer avatar. This result agrees with our qualitative observation of the images as well.



\subsection{Ablation Analysis}

In this section, we present an ablation analysis on several different components in the design choice of our system. The results are shown in Fig.~\ref{fig:ablation}.

First, we analyze our design of VAE (Sec.~\ref{sec:clothdecoder}) + temporal modeling (Sec. \ref{sec:animation}) for clothing animation. One alternative for this design is to combine the functionality of these two networks into one: to train a decoder that takes a sequence of skeleton poses as input and predicts clothing geometry and texture as output. The result of this comparison is shown on the left of Fig.~\ref{fig:ablation}. Here, the baseline model produces blurry output around the logo on the T-shirt. Even a sequence of skeleton poses does not contain enough information to fully determine the clothing state. Therefore, similar to the analysis in \cite{bagautdinov2021driving}, directly training a regressor from the information-deficient input to final clothing output leads to underfitting to the data by the model. In contrast, in our proposed system, the VAE network can model different clothing states in detail with a generative latent space, while the temporal modeling network infers the most probable clothing state. In this way, our method can produce high-quality animation output with sharp detail.

Next, we demonstrate the influence of photometric texture alignment (Sec.~\ref{sec:textureAlignment}) on the final animation output. We compare the results generated by our full model, which is trained on registered body and clothing data with texture alignment, against a baseline model trained on data without texture alignment (output of Sec.~\ref{sec:dataProcessing}). The result is shown in the middle of Fig.~\ref{fig:ablation}. We see that photometric texture alignment also helps to produce sharper detail in the animation output, as the better texture alignment makes the data easier for the network to model.

In addition, we also validate the ability of our network to generate view-dependent effects. We compare our full model with a baseline model where the body and clothing networks do not take view conditioning as input. The results are shown on the right of Fig.~\ref{fig:ablation}. Our output with view-dependent effects is visually more similar to the ground truth than the baseline model without view conditioning. The most obvious difference is observed near the silhouette of the subject, where the view-dependent output is brighter due to Fresnel reflectance when the incidence angle gets close to $90^\circ$ \cite{lafortune1997non}, an important factor that makes the view-dependent output more photo-realistic.

In the supplementary material, we also include a video comparison of animation results with different lengths $L$ of the temporal window as input to our TCN (Sec.~\ref{sec:animation}). With a small temporal window (for example $L = 1,3,8$), the temporal model tends to produce output with jittering. We find $L = 15$ or $30$ achieves a good tradeoff between visual temporal consistency and model efficiency. For a more detailed analysis of this issue, please refer to the supplementary document.

\subsection{Application: Clothing Texture Editing}

In this section, we demonstrate editing for the clothing texture. On top of our photorealistic animation output, we further edit the clothing pattern in four different styles. First, we multiply the RGB channels of the clothing UV texture with different scaling factors to modify the color of the clothing. Second, we apply a checkerboard pattern on our clothing layer. Third, we ask an artist to create a stylistic pattern and then apply it to our clothing animation output. Fourth, we add the ACM SIGGRAPH Logo and text to the front side of the clothing. The results are shown in Fig.~\ref{fig:texture-editing}. Once the desired pattern is determined, our model can produce animation with the edited texture for any motion sequence similar to those shown in Sec.~\ref{sec:eval-animation}.

Compared with the single-layer model, our two-layer structure naturally allows us to easily manipulate the clothing texture in the UV space without interfering with the inner layer in a temporally coherent manner. For comparison, we apply the same blue color transformation to the single-layer output. For this purpose, we manually segment out the clothing region for the first frame in the sequence in the UV space, and apply the color transformation in the segmented region to all the following frames. This approach produces reasonable results for the first frame (shown on the first column of Fig.~\ref{fig:texture-editing-compare-single}); for the following frames, however, applying the color transformation in the same UV region will suffer from misalignment of the edited area and actual clothing region, as shown in the right two columns of Fig.~\ref{fig:texture-editing-compare-single}. The visual artifact caused by this misalignment is highlighted in the zoomed-in boxes in the figure.

%% file: conclusion.tex
\section{Discussion}
We have proposed a two-layer mesh representation for building an animatable avatar for clothed body. Results have demonstrated that the explicit clothing modeling not only improves the rendered clothing quality in animation, but also enables the editability of the clothing texture, opening up new possibilities for codec avatars. The two-layer avatar models cannot be obtained without the success of two-layer registration of the clothed body. We thus have presented a new clothed body registration method along with a texture alignment method to improve the photometric correspondences using inverse rendering.


Our clothed body model is trained for each individual subject and also can only be animated for that individual. All the driving signals have been captured from the same subject performing social interactions. The animatable model may not be able to generalize to poses deviating significantly from the training pose distribution. Artifacts may appear if our model is used for arbitrary motion retargeting.

In this work, we are only focusing on T-shirts. To extend the work to lower body clothing, like short pants with the boundary shifting on the legs, we need to extend the current two-layer work to handle multiple layers, potentially with occlusion between layers, which poses additional challenges to both registration and modeling. Another common piece of clothing is a skirt, which could be even more difficult due to its large motion and deformation. We cannot handle topology-changing clothing, like opening a zipped jacket.

Even with the current two-layer framework, our clothing registration method would fail if the hands and clothing interact significantly, for example, hands dragging the clothing or hands put under the clothing. The current non-physical interaction modeling between clothing and body may not easily extend to handle these challenges. One possibility is to integrate more physical constraints into registration and learning for animation. 